\documentclass{article}
\usepackage{graphicx} 
\usepackage{amsmath}
\usepackage{amssymb}
\usepackage{appendix}
\usepackage{hyperref}
\usepackage{mathtools}
\usepackage{amsthm}
\usepackage{tikz}
\usepackage{float}
\usepackage{hyperref}
\usetikzlibrary{fit}
\usetikzlibrary{positioning}
\usepackage{xcolor}
\theoremstyle{plain}

\theoremstyle{definition}

\theoremstyle{remark}

\usepackage[accepted]{icml2024}
\begin{document}

\twocolumn[
\icmltitle{zELO: ELO-inspired Training Method for Rerankers and Embedding Models}
\icmlsetsymbol{equal}{*}

\begin{icmlauthorlist}
\icmlauthor{Nicholas Pipitone}{yyy}
\icmlauthor{Ghita Houir Alami}{yyy}
\icmlauthor{Advaith Avadhanam}{yyy}
\icmlauthor{Anton Kaminskyi}{yyy}
\icmlauthor{Ashley Khoo}{yyy}
\end{icmlauthorlist}

\icmlaffiliation{yyy}{ZeroEntropy, San Francisco, CA}

\icmlcorrespondingauthor{Nicholas Pipitone}{npip99@gmail.com}
\icmlcorrespondingauthor{Ghita Houir Alami}{ghita.houir-alami@polytechnique.edu}
\icmlcorrespondingauthor{Advaith Avadhanam}{advaith.ava@gmail.com}
\icmlcorrespondingauthor{Anton Kaminskyi}{}
\icmlcorrespondingauthor{Ashley Khoo}{}
\vskip 0.3in
]

\section{Abstract}

We introduce a novel training methodology named \textbf{zELO}, which optimizes retrieval performance via the analysis that ranking tasks are statically equivalent to a Thurstone model. Based on the zELO method, we use unsupervised data in order train a suite of state-of-the-art open-weight reranker models: \textbf{zerank-1} and \textbf{zerank-1-small}. These models achieve the highest retrieval scores in multiple domains, including finance, legal, code, and STEM, outperforming closed-source proprietary rerankers on both NDCG@10 and Recall. These models also demonstrate great versatility, maintaining their 0-shot performance on out-of-domain and private customer datasets. The training data included $112,000$ queries and $100$ documents per query, and was trained end-to-end from unannotated queries and documents in less than 10,000 H100-hours.

\section{Summary of Contributions}

\subsection{\textbf{zELO}: A novel Elo-based multi-stage training pipeline}

We introduce a novel multi-stage training process inspired by Elo scoring systems. First, we generate candidate documents using a first-stage retriever (e.g. ZeroEntropy’s Search API, or Lucene + Embedding hybrid search). Then, we gather sparse pairwise preferences from an ensemble of large language models (and, for scale, a pairwise SLM distilled from the ensemble of LLMs). These pairwise preferences are converted into absolute relevance scores using the Thurstone statistical model. Finally, we fine-tune our pointwise rerankers on these query-document \textbf{zELO} scores.

\subsection{Open-source reranker models released}
We release two fully open-weight rerankers trained on the \textbf{zELO} Method: \textbf{zerank-1}, initialized from Qwen3-4B (Yang et al. 2025), and \textbf{zerank-1-small}, initialized from Qwen3-1.7B (Yang et al. 2025). Rerankers are cross-encoder models that take a query-document pair as input and output a relevance score between 0 and 1, significantly boosting the performance of first-stage search methods such as BM25, embedding search, and hybrid retrieval. Both models were trained on 112,000 query–corpus pairs, i.e. over 5 million query-zELO pairs, and their weights are available on Huggingface. zerank-1-small in particular, is released under the permissible Apache 2.0 License. zerank-1 is open-weight and available for license by the ZeroEntropy team.

\textbf{zerank-1}: \url{https://huggingface.co/zeroentropy/zerank-1}

\textbf{zerank-1-small}: \url{https://huggingface.co/zeroentropy/zerank-1-small}

\subsection{State-of-the-art reranking accuracy across domains, languages, and retrieval methods}

\textbf{zerank-1} consistently outperforms commercial rerankers twice its size, with NDCG improvements of up to 5 points on benchmarks in finance, medicine, legal, code, and math. It also outperforms much larger LLMs-as-a-reranker, such as Gemini Flash 2.0, GPT-4o-mini, and GPT-5-mini/nano. It delivers strong gains regardless of the initial retrieval method, including BM25, embedding-based, and hybrid search.

\subsection{Unsupervised training and fine-tuning}

Most importantly, at ZeroEntropy we found experimentally that ensembles of LLMs via \textbf{zELO} generate \textbf{higher quality data than an equivalent number of human annotators} on average. The zELO method has strong convergence properties. We scale ensemble inferences until the zELO score of the target document converges, which gives a strong indicator of fundamental query-document relevancy. The entire annotation method has been open sourced by ZeroEntropy as zbench (https://github.com/zeroentropy-ai/zbench). zELO can be used both for benchmarking internal private documents, and for generating domain-specific fine-tuning data.

Because zELO is fully automated, it can also be used for \textbf{live production evaluations}. While a given reranker is used in production, live query logs can be randomly sampled, annotated automatically via zELO, and can be used to easily discover and fix issues in a live production retrieval pipeline (For example: if necessary context was never ingested, or if the initial retriever couldn't find it, etc). The annotations can also be used to fine-tune the reranker live, or if given per-customer context, can be used for personalized recommendation systems.

\section{Motivation}

\subsection{Existing SOTA}

Rerankers are the most fundamental operation in information retrieval. They take in a query document pair $(q, d)$, and output a score $s \in [0, 1]$. If it was feasible, every query into a search engine would simply have a reranker do a linear scan over the entire corpus. However, because this is computationally infeasible when the corpus is large, vector-based approximators are used for selecting initial candidates. These include sparse lexical keyword search such as \textbf{BM25} (Robertson et al. 1995), along with transformer-based \textbf{embedding} models. Keyword search fails if the user fails to recall the exact keyword, and dense embedding models can't converge on true relevancy because of the limitations of N-dimensional space (\href{https://arxiv.org/pdf/2508.21038}{Weller et al., 2025}).

Given the fundamental importance of rerankers, we note that rerankers can always learn from an SFT distillation task. We have some "teacher" reranker, traditionally human-annotated binary relevance labels, and then a "student" reranker, often an ML model, we can train the student on the teacher annotations.

The current SOTA method for training rerankers involves InfoNCE triplet loss $(q, d^+, d^-)$. Humans (the teacher) are tasked with annotated the positive, while the sampling method for negatives remains a free parameter. The most basic negative sampling strategy is in-batch random sampling (E.g. CLIP). However, the signal from random sampling is extremely weak, given that the sampled $d^-$ is usually obviously irrelevant; this method thus requires extremely large batch size ($>100k$), which in turn forces the use of poor quality web-scale data. The SOTA training method is "hard negative mining", wherein there is an attempt to make $d^-$ as relevant as possible to maximize the signal from contrastive learning -- often via an ensemble of embedding models followed by an ensemble of rerankers (Even LLM-as-a-reranker).

\subsection{Laffer Curve: The Fundamental Constraint on Existing SOTA Hard Negative Mining}
\label{sec:laffer}

Experimentally, we found that by making the "hard negative miner" as intelligent as possible, the Model eventually did not learn any better. In fact, we found that it got \textit{significantly worse}. Manual inspection made the issue clear: The hard negatives were on average legitimately more relevant than the human-annotated positive. This is unavoidable, as humans cannot exhaustively scan an entire corpus, and SOTA methods such as LLM-ensemble rerankers can reason on a much larger knowledge base than even expert annotators -- and do so at scale.

While one could human annotate $(q, d^-, d^+)$ to confirm $d^-$ as a true negative vis a vis the positive, this is inherently a pairwise comparison. For a pointwise model, absolute scoring via InfoNCE requires in-batch negatives, which requires an unsupervised negative sampling strategy. Thus, the intractability of false negatives in hard negative mining remains.

\begin{figure}
    \centering
    \includegraphics[width=1\linewidth]{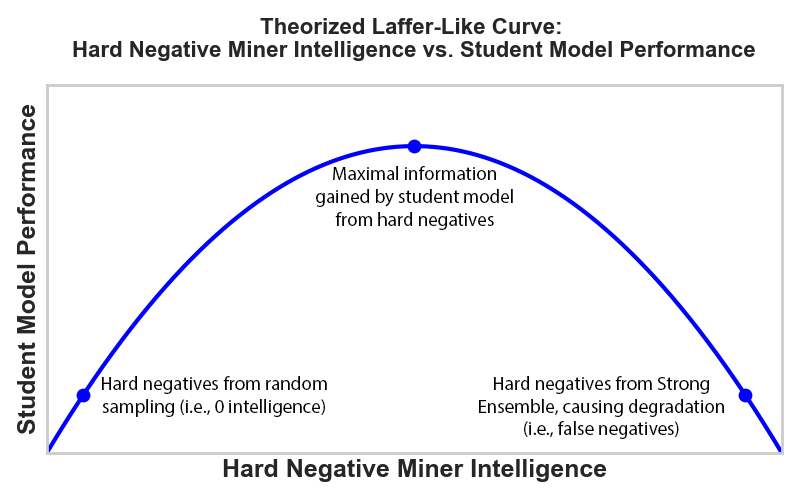}
    \caption{The proposed Laffer curve between hard negative miner intelligence and student model performance. We did not collect sufficient data with miner intelligence as the only independent variable, so we leave such data collection to future research.}
    \label{fig:enter-label}
\end{figure}

We conclude that hard negative mining techniques create a Laffer curve with respect to hard negative quality: As ensemble-generated hard negatives approach and exceed the quality of human-positive annotations, the marginal benefit from the distillation process diminishes and eventually becomes negative, indicating a fundamental limitation in the hard negative mining methodology.

From here, we make the following argument: the highest possible pointwise reranker performance is not that which corresponds to the optimal point on this induced Laffer curve. "Hard negative mining" is fundamentally flawed and its resulting accuracy is fundamentally capped by the training algorithm itself. \textbf{This was the motivating reason for using pairwise annotations generated by ensembles of frontier models and their resulting ELO scores as our source of ground truth, as well as for the development of a novel training method which has no theoretical limitations on student model performance other than the accuracy of the teacher model itself.}

\section{The zELO Method}
\label{sec:zELO}
\subsection{Definitions:}
Define $\mathcal{Q}$ to be the space of all queries, and $\mathcal{D}$ to be the space of all documents.

In this formalism, a pointwise reranker function $R_{\text{point}}$ is a function 
$$R_{\text{point}}: \mathcal{Q} \times \mathcal{D} \rightarrow [0, 1] \subset \mathbb{R}$$
Such that, for a given query $q\in\mathcal{Q}$ and corpus $C=\{d_1, \ldots, d_n\}\subset \mathcal{D}$ of documents, if $i_1, \ldots, i_n$ is the relevance ranking of documents, we have:
$$R_{\text{point}}(q, d_{i_1}) > R_{\text{point}}(q, d_{i_2}) > \ldots > R_{\text{point}}(q, d_{i_n})$$
Additionally, we may define a pairwise reranker to be a function
$$R_{\text{pair}}: \mathcal{Q} \times \mathcal{D} \times \mathcal{D} \rightarrow [0, 1]$$
Such that:
$$p_{ij}:=R_{\text{pair}}(q, d_i, d_j)$$
constitutes the probability that, w.r.t some query $q \in \mathcal{Q}$, document $d_i \in \mathcal{D}$ is preferred over document $d_j \in \mathcal{D}$. Or, equivalently, such that $R_{\text{pair}}(q, d_i, d_j) -\frac{1}{2}$ is the strength by which $d_i$ is \textit{more} relevant than $d_j$  (consider $R_{\text{pair}}(q, d_i, d_i) = \frac{1}{2}$ indicates no preference between two identical documents). Ideally, $R_{\text{pair}}$ induces a total order on $\mathcal{D}$ for every $q$, but this is not strictly required.
\subsection{Using $R_{\text{pair}}$ to generate SFT}
As discussed in Section 3.1, for any given query, rerankers generally operate on a subset of the top $k$ documents $\mathcal{Z} \subset \mathcal{D}$ (where $|\mathcal{Z}|=k$) given by some initial retrieval method (BM-25, Embeddings, Hybrid, etc.) Typical values for $k$ are $k\leq 100$. Furthermore, given that pointwise rerankers constitute the models actually used for IR tasks, we first train the pairwise $R_{\text{pair}}$ with the intention of using this to create a SFT dataset upon which to train $R_{\text{point}}$. 

Specifically, for a fixed query $q$ with $k$ many associated retrieved documents, inference $R_{\text{pair}}$ on pairs of documents $d_i,d_j\in Z$ given subset $\mathcal{Z}\subset C$ to obtain scores $p_{i,j}$. Overall, we form a dense preference matrix $P$ that is $k \times k$:

$$P = \begin{pmatrix}
p_{11} & p_{12} & p_{13} & \cdots & p_{1k} \\
p_{21} & p_{22} & p_{23} & \cdots & p_{2k} \\
p_{31} & p_{32} & p_{33} & \cdots & p_{3k} \\
\vdots & \vdots & \vdots & \ddots & \vdots \\
p_{k1} & p_{k2} & p_{k3} & \cdots & p_{kk}
\end{pmatrix}$$

Where the antisymmetry constraint $p_{ji} = 1 - p_{ij}$ ensures that:

$$P + P^T = \mathbf{1}\mathbf{1}^T$$

Where $\mathbf{1}$ is the $n$-dimensional vector of ones, where diagonal elements then satisfy $p_{ii} = \frac{1}{2}$.

Now define the sparse $n \times n$ matrix $W$ as the extension of the preference matrix $P$ to the full corpus $C$, where non-inferred pairs are set to zero:

$$w_{ij} = \begin{cases}
P_{ij} = R_{\text{pair}}(q, d_i, d_j) & \text{if } (d_i, d_j) \text{ was inferred} \\
0 & \text{if } (d_i, d_j) \text{ was not inferred}
\end{cases}$$

More formally, let $\mathcal{I} \subseteq [n] \times [n]$ denote the set of index pairs $(i,j)$ for which pairwise comparisons were performed. Then:

$$W = \begin{pmatrix}
w_{11} & w_{12} & w_{13} & \cdots & w_{1n} \\
w_{21} & w_{22} & w_{23} & \cdots & w_{2n} \\
w_{31} & w_{32} & w_{33} & \cdots & w_{3n} \\
\vdots & \vdots & \vdots & \ddots & \vdots \\
w_{n1} & w_{n2} & w_{n3} & \cdots & w_{nn}
\end{pmatrix}$$

where each entry satisfies:
\begin{enumerate}
    \item $w_{ij} + w_{ji} = 1$ if $(i,j) \in \mathcal{I}$ (inferred pairs maintain antisymmetry)
    \item $w_{ij} = w_{ji} = 0$ if $(i,j) \notin \mathcal{I}$ (non-inferred pairs are zero)
\end{enumerate}

The transformation from these pairwise preferences to pointwise scores follows the Bradley-Terry model framework. Given pairwise comparison matrix $W$, we seek to find latent "abilities" (in the language of BT) or relevance scores (in that of IR) that explain the observed preferences.
\subsubsection{Connection to Bradley-Terry Model}

The Bradley-Terry model assumes that for documents $d_i$ and $d_j$ with latent abilities $\pi_i$ and $\pi_j$, the probability that $d_i$ is preferred over $d_j$ is:

$$\mathbb{P}(d_i \succ d_j) = \frac{\pi_i}{\pi_i + \pi_j}$$

In the Elo rating system formulation of Bradley-Terry, we parameterize $\pi_i = e^{\text{Elo}_i}$ where $\text{Elo}_i$ is the Elo rating of the document $d_i$, giving:

$$\mathbb{P}(d_i \succ d_j) = \frac{e^{\text{Elo}_i}}{e^{\text{Elo}_i} + e^{\text{Elo}_j}} = \frac{1}{1 + e^{-(\text{Elo}_i - \text{Elo}_j)}}$$
$$ = \sigma(\text{Elo}_i - \text{Elo}_j)$$

As such, we fit $\{\text{Elo}_1, \text{Elo}_2, ..., \text{Elo}_n\}$ such that whenever $w_{ij}\neq 0$:
$$w_{ij} = \sigma(\text{Elo}_i - \text{Elo}_j)$$
And secondarily, for the purposes of normalization across queries, we constrain:
$$e_1 + e_2 + ... + e_n = 0$$
We define a negative log likelihood loss and fit Elos by gradient descent via maximum likelihood estimation. The loss is then
$$\mathcal{L} = -\sum_{i, j}{w_{ij}\log(\frac{p_{i}}{p_{i}+p_{j}})} = \sum_{i, j}{w_{ij}\log(1+e^{\text{Elo}_j-\text{Elo}_i})}$$
And it is a known result from Zermelo that this has a unique local minimum (given a final constraint like $\text{Elo}_1+\ldots+\text{Elo}_n = 0$) (Zermelo 1929).  As a result, gradient descent will converge should we have a time-decaying learning rate $\eta_t$ with $\sum_{t \geq 1}{\eta_t}$ diverging (we use $\eta_t = t^{-0.125}$ here).

\subsubsection{Thurstone Model}

The Thurstone Model posits that rankings are determined by per-document hidden scores (Which we call ELO), and that the probability that document $i$ is better than document $j$ is

$$w_{ij} = \frac{1+\text{erf}(\text{Elo}_i - \text{Elo}_j)}{2}$$

This is almost identical to Bradley-Terry, albeit assuming that a document's intrinsic noise takes a normal distribution rather than a gumbel distribution. We observe that this adjustment makes a better fit to the observed data, and is also well-justified (Justified via the central limit theorem, with the prior that the document comparison is subject to multiple sources of noise). In our final training run, we utilize Thurstone's $\text{erf}()$ rather than Bradley-Terry's $\sigma()$ for this reason.

\subsubsection{Extension to Plackett-Luce for Complete Rankings}

While Bradley-Terry handles pairwise comparisons, the  Plackett-Luce model generalizes this to complete rankings. For a ranking $\pi$ of documents $d_1, \ldots, d_n$, the Plackett-Luce probability is:
$$\mathbb{P}(\pi) = \prod_{i=1}^{n} \frac{\pi_{\pi(i)}}{\sum_{j=i}^{n} \pi_{\pi(j)}}$$
In our context, this allows us to model the probability of observing a complete ranking based on our Elo scores. However, since we work with sparse pairwise comparisons for computational tractability rather than complete rankings (more on this in section 5), we primarily consider the Bradley-Terry formulation.

\subsubsection{Sparse Matrix Subsampling for Elos}
\label{sec:elo_score_heuristics}

A dense $n \times n$ inference procedure for every query $q$ in the dataset is prohibitively expensive given the scale of $n$. Therefore, we sparsely infer $R_{\text{pair}}(q, d_i, d_j)$ for selected $O(n)$ $d_i, d_j$ pairs in a manner that the predicted Elos $e'_1, e'_2, \ldots, e'_n$ from the Elo calculation algorithm on the sparse matrix match the actual Elos $e_1, \ldots, e_n$ as well as possible. 

Let $G = (V, E)$ be our comparison graph where $V = \{d_1, \ldots, d_n\}$ and $(d_i, d_j) \in E$ iff $R_{\text{pair}}(q, d_i, d_j)$ is inferred. Let $\text{dist}_G(d_i, d_j)$ denote the shortest-path distance between vertices $d_i$ and $d_j$ in $G$, and let $\text{deg}(d_i)$ denote the degree of vertex $d_i$.

To ensure accurate Elo score estimation from sparse pairwise comparisons, our graph $G$ must satisfy three key structural properties:

\begin{figure}[H]
\centering
\textbf{1) Connected Graph:}

\

\begin{tikzpicture}[scale=0.75]
\node[circle,draw] (1) at (0,0) {$d_1$};
\node[circle,draw] (2) at (1,1) {$d_2$};
\node[circle,draw] (3) at (1,-1) {$d_3$};
\draw (1) -- (2);
\draw (1) -- (3);

\node[circle,draw] (4) at (3,-1) {$d_4$};
\node[circle,draw] (5) at (3,1) {$d_5$};
\draw (4) -- (5);

\node at (1.5,-2) {\textcolor{red}{Disconnected}};

\node[draw, rounded corners, fit={(1) (2) (3) (4) (5) (2,-2)}, inner sep=8pt, gray, dashed] {};
\end{tikzpicture}
\hspace{0.5cm}
\begin{tikzpicture}[scale=0.75]
\node[circle,draw] (1) at (0,0) {$d_1$};
\node[circle,draw] (2) at (1,1) {$d_2$};
\node[circle,draw] (3) at (1,-1) {$d_3$};
\node[circle,draw] (4) at (3,-1) {$d_4$};
\node[circle,draw] (5) at (3,1) {$d_5$};
\draw (1) -- (2);
\draw (1) -- (3);
\draw (2) -- (4);
\draw (3) -- (4);
\draw (4) -- (5);

\node at (1.5,-2) {\textcolor{green}{Connected}};

\node[draw, rounded corners, fit={(1) (2) (3) (4) (5) (2,-2)}, inner sep=8pt, gray, dashed] {};
\end{tikzpicture}
\caption{Left: Disconnected graph cannot establish relative Elo relationships between any of $\{d_1,d_2,d_3\}$ and any of $\{d_4,d_5\}$. Right: Connected graph enables global Elo ranking.}
\end{figure}
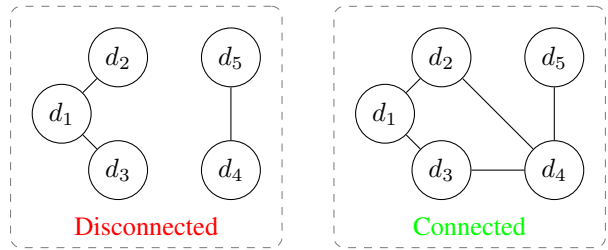
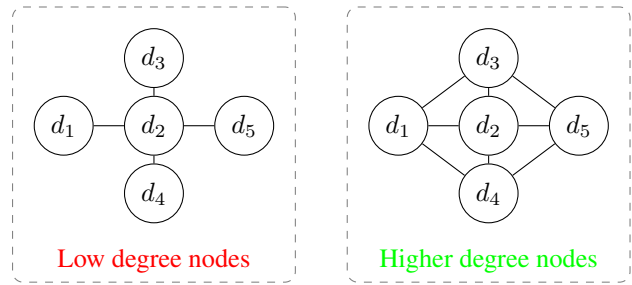
\begin{figure}[H]
\centering
\textbf{2) No Nodes With Low Degree:}

\

\begin{tikzpicture}[scale=0.6]
\node[circle,draw] (1) at (0,0) {$d_1$};
\node[circle,draw] (2) at (2,0) {$d_2$};
\node[circle,draw] (3) at (2,1.5) {$d_3$};
\node[circle,draw] (4) at (2,-1.5) {$d_4$};
\node[circle,draw] (5) at (4,0) {$d_5$};
\draw (1) -- (2);
\draw (2) -- (3);
\draw (2) -- (4);
\draw (2) -- (5);

\node at (2,-3) {\textcolor{red}{Low degree nodes}};

\node[draw, rounded corners, fit={(1) (2) (3) (4) (5) (1.5,-3)}, inner sep=8pt, gray, dashed] {};
\end{tikzpicture}
\hspace{0.5cm}
\begin{tikzpicture}[scale=0.6]
\node[circle,draw] (1) at (0,0) {$d_1$};
\node[circle,draw] (2) at (2,0) {$d_2$};
\node[circle,draw] (3) at (2,1.5) {$d_3$};
\node[circle,draw] (4) at (2,-1.5) {$d_4$};
\node[circle,draw] (5) at (4,0) {$d_5$};
\draw (1) -- (2);
\draw (2) -- (3);
\draw (2) -- (4);
\draw (2) -- (5);
\draw (1) -- (3);
\draw (1) -- (4);
\draw (3) -- (5);
\draw (4) -- (5);

\node at (2,-3) {\textcolor{green}{Higher degree nodes}};

\node[draw, rounded corners, fit={(1) (2) (3) (4) (5) (1.5,-3)}, inner sep=8pt, gray, dashed] {};
\end{tikzpicture}
\caption{Left: Node $d_1$ has degree 1, making its Elo estimate unreliable. Right: All nodes have degree $\geq 3$, providing more stable estimates.}
\end{figure}
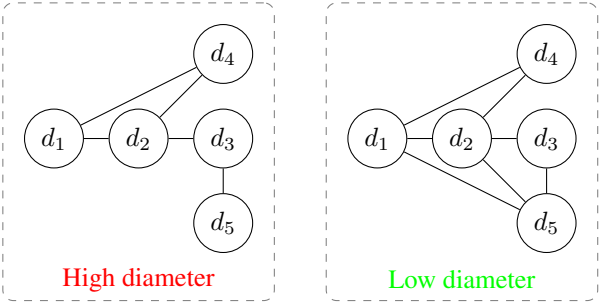
\begin{figure}[H]
\centering
\textbf{3) Low Diameter:}

\

\begin{tikzpicture}[scale=0.75]
\node[circle,draw] (1) at (0,0) {$d_1$};
\node[circle,draw] (2) at (1.5,0) {$d_2$};
\node[circle,draw] (3) at (3,0) {$d_3$};
\node[circle,draw] (4) at (3,1.5) {$d_4$};
\node[circle,draw] (5) at (3,-1.5) {$d_5$};
\draw (1) -- (2);
\draw (1) -- (4);
\draw (2) -- (3);
\draw (4) -- (2);
\draw (3) -- (5);

\node at (1.5,-2.5) {\textcolor{red}{High diameter}};

\node[draw, rounded corners, fit={(1) (2) (3) (4) (5) (1.5,-2.5)}, inner sep=8pt, gray, dashed] {};
\end{tikzpicture}
\hspace{0.5cm}
\begin{tikzpicture}[scale=0.75]
\node[circle,draw] (1) at (0,0) {$d_1$};
\node[circle,draw] (2) at (1.5,0) {$d_2$};
\node[circle,draw] (3) at (3,0) {$d_3$};
\node[circle,draw] (4) at (3,1.5) {$d_4$};
\node[circle,draw] (5) at (3,-1.5) {$d_5$};
\draw (1) -- (2);
\draw (1) -- (5);
\draw (2) -- (3);
\draw (1) -- (4);
\draw (2) -- (4);
\draw (2) -- (5);
\draw (3) -- (5);
\node at (1.5,-2.5) {\textcolor{green}{Low diameter}};

\node[draw, rounded corners, fit={(1) (2) (3) (4) (5) (2.5,-2.5)}, inner sep=8pt, gray, dashed] {};
\end{tikzpicture}
\caption{Left: Documents $d_1$ and $d_5$ are separated by 3 edges in this pendant structure, leading to high uncertainty in their relative Elo. Right: Maximum separation is 2 edges, providing more reliable comparisons.}
\end{figure}
In particular, we employ the following heuristics to guide our sparse subsampling strategy:

\begin{enumerate}
    \item \textit{Path Length Heuristic}: The variance in estimated Elo differences is proportional to graph distance:
    $$\text{Var}[e'_i - e'_j] \propto \text{dist}_G(d_i, d_j)$$
    
    \item \textit{Degree Stability Heuristic}: The variance of individual Elo estimates is inversely proportional to node degree:
    $$\text{Var}[e'_i] \propto \frac{1}{\text{deg}(d_i)}$$
\end{enumerate}

These heuristics motivate our graph construction constraints:
$$\text{diam}(G) = \max_{i,j} \text{dist}_G(d_i, d_j) \leq \rho$$
$$\min_i \text{deg}(d_i) \geq \delta$$

Under these heuristics, the probability of ranking error between documents $d_i$ and $d_j$ increases with their graph distance and decreases with the true Elo gap $|e_i - e_j|$.

Now, with $O(n)$ edges, the combination of these two constraints naturally lends itself to considering $k$-regular graphs. Consider that it is a known theoretical result that for a random $k$-regular graph $G$:
$$\text{diam}(G) \leq \log_{k-1}(n)+\log_{k-1}(\log(n))+\log_{k-1}(\frac{5}{2}k(k-1))$$
With probability asymptotically 1, as per Bollobás.\footnote{(Actually, $\frac{5}{2}$ may be replaced with $2+\epsilon$ for any $\epsilon > 0$) (Bollobás 2001, Chapter 10.3)}

Generating arbitrary random $k$-regular graphs is not efficient, however, and to ensure good connectivity, it is actually better to choose $\frac{k}{2}$ random $n$-cycles and union their edge sets over the vertices; the resulting graph is $k$-connected (and if no $(d_i,d_j)$ are duplicated across cycles, the graph is in fact $k$-regular as well). What is more, such a graph can be generated easily in $O(n)$ time by taking random permutations. This graph will have $N = \frac{kn}{2}$ edges. Below we illustrate a toy example with this method.
\begin{figure}[H]
\centering
\textbf{Constructing a $k$-regular graph via cycle splicing}
\textbf{(example: $n = 6$, $k = 4$)}

\vspace{0.3cm}

\textbf{Step 1: Generate $k/2 = 2$ Random Cycles}
\begin{tikzpicture}[scale=0.7]
\vspace{0.3cm}
\begin{scope}[xshift=0cm]
\node[circle,draw,fill=blue!20] (1) at (0,1.5) {$d_1$};
\node[circle,draw,fill=blue!20] (2) at (1.3,0.7) {$d_2$};
\node[circle,draw,fill=blue!20] (3) at (1.3,-0.7) {$d_3$};
\node[circle,draw,fill=blue!20] (4) at (0,-1.5) {$d_4$};
\node[circle,draw,fill=blue!20] (5) at (-1.3,-0.7) {$d_5$};
\node[circle,draw,fill=blue!20] (6) at (-1.3,0.7) {$d_6$};

\draw[blue,thick,->] (1) -- (2);
\draw[blue,thick,->] (2) -- (3);
\draw[blue,thick,->] (3) -- (4);
\draw[blue,thick,->] (4) -- (5);
\draw[blue,thick,->] (5) -- (6);
\draw[blue,thick,->] (6) -- (1);
\end{scope}

\begin{scope}[xshift=5cm]
\node[circle,draw,fill=red!20] (1) at (0,1.5) {$d_1$};
\node[circle,draw,fill=red!20] (2) at (1.3,0.7) {$d_2$};
\node[circle,draw,fill=red!20] (3) at (1.3,-0.7) {$d_3$};
\node[circle,draw,fill=red!20] (4) at (0,-1.5) {$d_4$};
\node[circle,draw,fill=red!20] (5) at (-1.3,-0.7) {$d_5$};
\node[circle,draw,fill=red!20] (6) at (-1.3,0.7) {$d_6$};

\draw[red,thick,->] (1) -- (5);
\draw[red,thick,->] (5) -- (2);
\draw[red,thick,->] (2) -- (6);
\draw[red,thick,->] (6) -- (3);
\draw[red,thick,->] (3) -- (4);
\draw[red,thick,->] (4) -- (1);
\end{scope}
\end{tikzpicture}

\vspace{0.2cm}
{\small \textcolor{blue}{Cycle 1: $1 \to 2 \to 3 \to 4 \to 5 \to 6 \to 1$}}
\vspace{0.2cm}
{\small \textcolor{red}{\\Cycle 2: $1 \to 5 \to 2 \to 6 \to 3 \to 4 \to 1$}}

\caption{Generate two random cycles over the same vertex set.}
\label{fig:randomcycles}
\end{figure}

\begin{figure}[H]
\centering
\vspace{0.4cm}

\vspace{0.5cm}
\textbf{Step 2: Overlay the Cycles}

\begin{tikzpicture}[scale=0.8]
\node[circle,draw,fill=purple!20] (1) at (0,2) {$d_1$};
\node[circle,draw,fill=purple!20] (2) at (1.7,1) {$d_2$};
\node[circle,draw,fill=purple!20] (3) at (1.7,-1) {$d_3$};
\node[circle,draw,fill=purple!20] (4) at (0,-2) {$d_4$};
\node[circle,draw,fill=purple!20] (5) at (-1.7,-1) {$d_5$};
\node[circle,draw,fill=purple!20] (6) at (-1.7,1) {$d_6$};

\draw[blue,thick] (1) -- (2);
\draw[blue,thick] (2) -- (3);
\draw[blue,thick] (3) -- (4);
\draw[blue,thick] (4) -- (5);
\draw[blue,thick] (5) -- (6);
\draw[blue,thick] (6) -- (1);

\draw[red,thick] (1) -- (5);
\draw[red,thick] (5) -- (2);
\draw[red,thick] (2) -- (6);
\draw[red,thick] (6) -- (3);
\draw[red,thick] (3) -- (4);
\draw[red,thick] (4) -- (1);
\end{tikzpicture}

\vspace{0.3cm}
\textbf{Degree of each node ($k=4$):} \\
$\deg(d_1) = 4$ \quad $\deg(d_2) = 4$ \quad $\deg(d_3) = 4$ \quad $\deg(d_4) = 4$ \quad $\deg(d_5) = 4$ \quad $\deg(d_6) = 4$

\caption{Construction of a 4-regular graph by overlaying 2 random cycles. Total edges: $N = \frac{kn}{2} = \frac{4 \cdot 6}{2} = 12$.}
\end{figure}
\begin{figure}[H]
\centering
\textbf{Properties of the resulting graph}

\begin{tikzpicture}[scale=1.0]
\node[circle,draw,fill=purple!20] (1) at (0,2) {$d_1$};
\node[circle,draw,fill=purple!20] (2) at (1.7,1) {$d_2$};
\node[circle,draw,fill=purple!20] (3) at (1.7,-1) {$d_3$};
\node[circle,draw,fill=purple!20] (4) at (0,-2) {$d_4$};
\node[circle,draw,fill=purple!20] (5) at (-1.7,-1) {$d_5$};
\node[circle,draw,fill=purple!20] (6) at (-1.7,1) {$d_6$};

\draw[thick] (1) -- (2);
\draw[thick] (2) -- (3);
\draw[thick] (3) -- (4);
\draw[thick] (4) -- (5);
\draw[thick] (5) -- (6);
\draw[thick] (6) -- (1);
\draw[thick] (1) -- (5);
\draw[thick] (5) -- (2);
\draw[thick] (2) -- (6);
\draw[thick] (6) -- (3);
\draw[thick] (3) -- (4);
\draw[thick] (4) -- (1);

\end{tikzpicture}

\vspace{0.3cm}
\textbf{Low diam:} $\text{dist}(d_1, d_3) = 2$, $\text{diam}(G) = 2$ \quad 
\\\textbf{High connectivity:} Graph is 2-connected \quad 
\\\textbf{Uniform degree:} All nodes have degree 4

\caption{The resulting 4-regular graph has low diameter, high connectivity, and uniform degree distribution - ideal properties for sparse ELO estimation.}
\end{figure}
\subsection{Training $R_{\text{point}}$ from SFT}

Given a dataset of query-document pairs with relevance scores $\{(q_i, d_j, y_{ij})\}$ where $y_{ij} \in [0,1]$ represents the relevance of document $d_j$ to query $q_i$, we train the pointwise reranker $R_{\text{point}}$ using supervised fine-tuning.

We minimize the mean squared error loss:
$$\mathcal{L}_{\text{SFT}} = \frac{1}{|D_{\text{train}}|} \sum_{(q,d,y) \in D_{\text{train}}} (R_{\text{point}}(q,d) - y)^2$$

where $D_{\text{train}}$ is our training dataset.

Little needs to be said here; our procedure does not deviate meaningfully from standard practice other than that our $\{y_{ij}\}$ are given by the estimated $\{e'_{j}\}$ for each fixed query $q_i$ from the prior section as opposed to human-annotated binary scores.
\section{Training the Pairwise Reranker}

\subsection{Dataset}

As input to the \textbf{zELO} algorithm, we must collect $\mathcal{Q}, \mathcal{D}, \mathcal{Z}$. The $\mathcal{Q}, \mathcal{D}$ used by our final model consists of $112,000$ publicly available queries across a wide array of domains including finance, law, medicine, code, and STEM; and $>100M$ publicly available web-scale documents. For generating $\mathcal{Z}$, the initial retrieval method we use is cosine similarity on embeddings from $\textbf{Qwen3-Embedding-4B}$ (\href{https://arxiv.org/pdf/2506.05176}{Zhang et al., 2025}), combined via RRF with a lexical sparse BM25. For lexical sparse embeddings, we use an optimized multilingual tokenizer pipeline (language detection, followed by language-specific stemming). The top-$k$ chosen is $k=100$. Initial retrieval is done via the \href{https://docs.zeroentropy.dev/api-reference/queries/top-documents}{ZeroEntropy Search Engine}, using the endpoint \texttt{/queries/top-documents}.

\subsection{Ensemble Annotation as Source of Truth}

Consider a population $\mathcal{P}$ of annotators. We define the ensemble ranking as

$$_{\mathcal{P}}R_{\text{pair}}(q, d_i, d_j) = \frac{1}{|\mathcal{P}|} \sum_{p \in \mathcal{P}} {_p}R_{\text{pair}}(q, d_i, d_j)$$

We create a small internal dataset across numerous verticals (Medicine, Law, Finance, Code, etc) to use as the Gold Standard. For each query, the documents are selected via randomly sampling two documents out of the Top $k$ by initial retrieval. $k$ is randomly sampled between $10$ and $100$ with an inverse distribution. In order to resolve $R_{\text{ensemble}}$ for a particular input, we sample annotators until the standard error of the mean approaches $0.1$.

As per \href{sec:laffer}{Section 3.2 Laffer Curve}, we observe the difficulties that accompany using human annotations as the source of ground truth. While sampling until convergence across all conceivable documents $\mathcal{D}$ for an entire querying population $\mathcal{P}_{\text{universal}}$ against all conceivable queries $\mathcal{Q}$ would (definitionally) match the target ideal reranker, human annotation is in practice expensive, noisy, and observed to be simply inferior to an ensemble of SOTA large language models. As such, we preferentially inference the latter to generate pairwise annotations.

However, as discussed later in \href{sec:rlhf}{Section 5.6 RLHF}, we do for each query make use of our dataset's highest ranked human-annotated document in a second training run of our pairwise reranker. Thus, we recapture signal that pure distillation from an LLM-ensemble teacher model was unable to impart on a first pass. In this way, we make use of those human annotations with the highest signal-to-noise ratio without fully shackling ourselves to them as ground truth.

\subsection{Ensemble of LLMs as Synthetic Data Generators}

Practically, consider $\mathcal{P}$ to be a set of frontier LLMs, each prompted against a given $(q, d_i, d_j)$ to output a chain-of-thought justification for and final judgment of document preference, and consider $_{\mathcal{P}}R_{\text{pair}}$ as before. For our purposes, we found $|\mathcal{P}|=3$ to be a good mix of economical, convergent, and accurate.

For each query $q$, we randomly sample a document pair $\{d_i, d_j\}\subset \mathcal{Z}$ and prompt each LLM to score the relative preference on $[-1,1]$,\footnote{In \href{sec:zELO}{Section 4.1 Definitions for the zELO Method} we give the codomain of $R_{\text{pair}}$ as $[0,1]$. Experimentally, prompting LLM's for a [-1,1] codomain gave more uniform results. We in any case transform this back into the expected codomain later. We would also flip the order of prompted documents half the time and take the negative result to control for left/right LLM bias.} where:
\begin{enumerate}
    \item $-1$ indicates strong preference for $d_i$
    \item $1$ indicates strong preference for $d_j$
    \item $0$ indicates no preference
\end{enumerate} We then clamp the raw scores to $\{-1, 0, 1\}$ and individually prompt-engineer each model to achieve roughly uniform response distribution. See \href{sec:pairwiseprompt}{the appendix} for the base prompt used.

Next, we average over $\mathcal{P}$ to get an \textit{ensemble score}, $p_{ij}$ (a real number in $[-1, 1]$ with $|\mathcal{P}|r\in\mathbb{Z}$). We then map the ensemble score into the expected $[0, 1]$ range by the following update rule:
$$_{\text{new}}p_{ij} = \frac{1-p_{ij}}{2}$$
That is, if the first document is preferred, the resulting score is near $1$, while the second document being preferred results in a score near $0$. This ensemble score is $\frac{1}{2|\mathcal{P}|}$ of an integer.

\begin{figure*}[ht]
    \centering
    \includegraphics[width=1\linewidth]{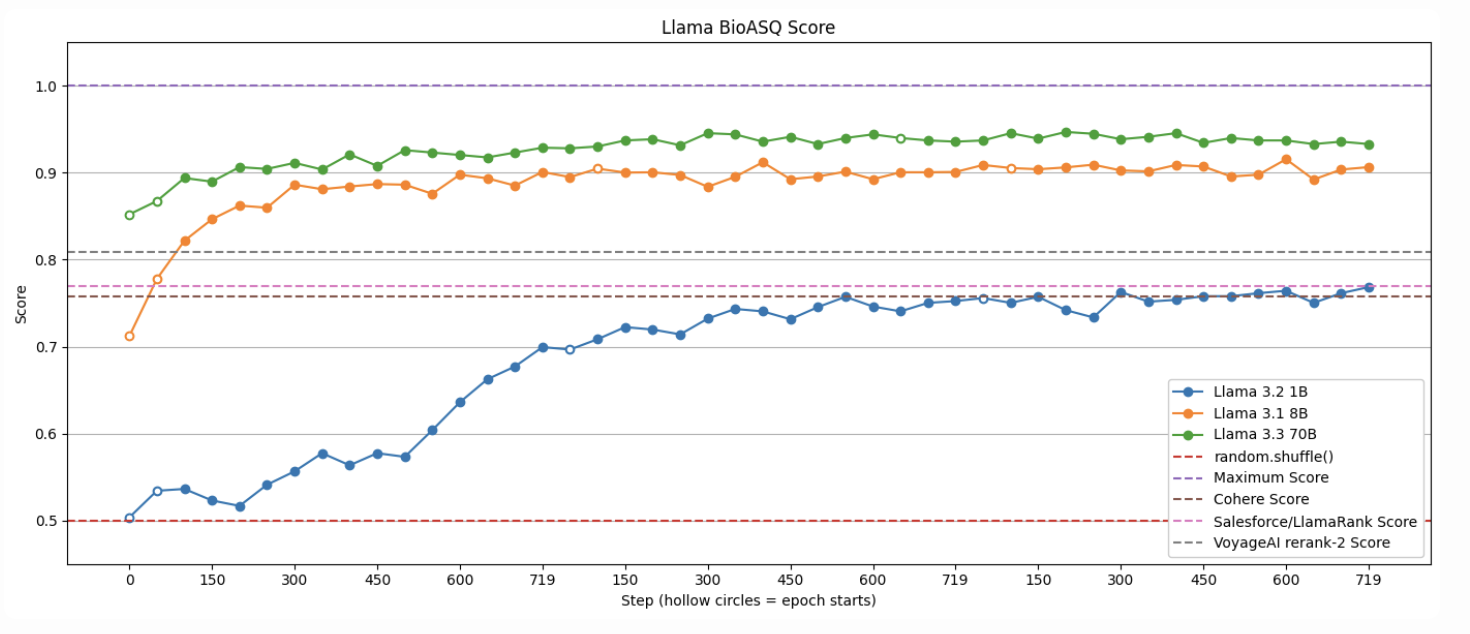}
    \caption{Frequency of the pairwise comparison models favoring documents on which the ensemble reaches consensus}
    \label{fig:training-run}
\end{figure*}

In contradistinction to $p_{ij}:=_{P}R_{\text{pair}}(q,d_i,d_j)$, the ensemble score, denote $p'_{ij}:=R'_{\text{pair}}(q, d_i, d_j)$ the score given by the pairwise reranker being trained to approximate our ground truth. We may simply use standard Binary Cross Entropy loss to train $R'_{\text{pair}}(q, d_i, d_j)$ on ensemble scores: $$\mathcal{L} = \sum_{i,j \text{ sampled over }q}{\text{BCE}(p_{ij},p'_{ij}})$$ 
Where, remember:
$$\text{BCE}(p, q) := -(p\log(q)+(1-p)\log(1-q))$$
Given that inferencing $_{P}R_{\text{pair}}$ is particularly expensive, we do so only once for each query, on a random $(d_i,d_j)$ pair taken from the top-$k$ initially-retrieved documents. Inferencing more pairs would in theory improve the performance of the student model, but given economic constraints, our limitation proved the number of high quality queries we managed to train over -- more entropy was to be had in inferencing over a new query than over an extra document pair for the same query.

We can see the convergence of this approach in Figure~\ref{fig:training-run} above. This graph is from an early research run which was not used to train our final model, but which demonstrates the generalizability of the approach across model sizes and datasets. Training three differently sized Llama distillates on BioASQ using this method and evaluating it against a validation subset of that dataset (with LLM-ensemble as source of ground truth, not human-annotations) showcases steady performance improvement and rapid convergence. In particular, we note that even for the 8B and 70B models with strong zero-shot performance, margins of 20\% and 10\% were noted, resulting in large performance gains vs SOTA rerankers (93\% vs 75\% to 81\% on this dataset). We further note that these models performed similarly on our internal standards of private data, implying generalization, whereas competing SOTA rerankers (admittedly pointwise) experienced a drop-off in performance, implying overfitting to evaluation datasets. 

\subsection{Elo Score Calculation}
Having trained the pairwise reranker $R'_{\text{pair}}$ and given a top-$k=100$ filtered corpus $\mathcal{Z}$ of $k$ documents, we can now infer it on some set $\mathcal{I}$ of pairs of documents $(d_i, d_j):i,j\in\mathcal{I}\subset [k]\times[k]$ for computational efficiency. We again swap the order $i,j$ half the time to mitigate the effects of potential biasing towards either position, and obtain scores $p_{ij}$, and set $p_{ji} = 1-p_{ij}$. We thus form the sparse matrix $P$ that is $k \times k$. 

This still leaves the choice of which $\mathcal{I}$ to infer $R'_{\text{pair}}$ on to generate $P$; this choice is important, as it determines the ELO estimates on which $R'_{\text{point}}$ will be trained. Recall from the section on \href{sec:elo_score_heuristics}{heuristics for subsampling sparse matrices for ELO scores} that we desire a few crucial characteristics of our sampling strategy:
\begin{enumerate}
    \item It must result in a connected graph, where every document is reachable from every other document by some chain of inferred pairwise comparisons.
    \item No document should be pairwise-inferred a very small number of times, lest its ELO prove unstable and erroneous. In particular, we determined that all documents should be inferred the same number of times.
    \item The maximum separation between two documents in the shortest-connecting chain of pairwise inferences (\textit{id est}, the diameter of the graph) should be low, otherwise relative ELO between them may prove unreliable (since pairwise document relevancy cannot be assumed to be strongly stochastically transitive).
\end{enumerate}
At ZeroEntropy, we explored various sampling methods to satisfy these constraints: simple random pairwise sampling without replacement, complete bipartite graphs $K_{l, k-l}$ (to get diameter of 2) and approximately $kl$ edges, as well as a dynamic method that selects successive $(d_i, d_j)$ pairs based on the smallest current ELO difference so as to maximize expected entropy. We evaluated these approaches by comparing ground-truth ELOs from densely inferred $k \times k$ matrices against ELOs generated from sparse sampling according to each method's graph structure.
\begin{figure}[H]
    \centering
    \includegraphics[width=\linewidth]{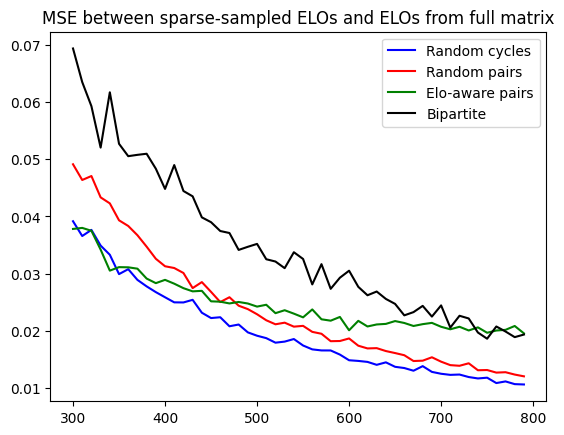}
    \caption{MSE between the predicted Elos and actual Elos, versus the number of "inferences" made to the matrix, for four sampling methods; random cycles, random pairs, the bipartite graph, and another method that calculates Elos live and favors picking $i, j$ with low current Elo absolute difference}
    \label{fig:mseloss}
\end{figure}
In the end, we discovered that a method of sampling $\frac{k}{2}$ random cycles (where $k$ is the desired valence of any document in a $k$-regular graph) and unioning the edge sets proved fastest in converging and generated the most stable results. See \hyperref[fig:randomcycles]{the illustration} of this process from Section 4.2.3 for more information. Results for these training procedure experiments are seen in Figure~\ref{fig:mseloss} above.

The convergence properties of the ELO ranking system are such that even with mere random sampling, the cross-entropy loss of ELO estimates from subsampling vs those calculated from the full $100\times 100$ matrix go to zero with only about a $1\%$ sampling ratio: 
\begin{figure}[H]
    \centering
    \includegraphics[width=0.9\linewidth]{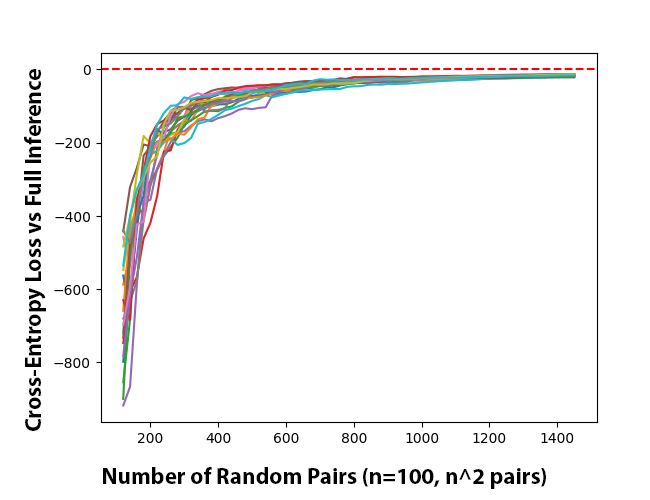} 
    \caption{Plots for a few different runs of random sampling; convergence observed around ~1.2k/10k total pairs sampled}
\end{figure}
However, we were able to further improve upon this with the method of cycles, ultimately using only $N = 400$ inferences ($0.4\%$ of full), with $k = 8$ ($4$ random cycles) to sample the Elos. These ELO scores become our pointwise estimates on which we train $R'_{\text{point}}$ in the next section.

\subsection{Training the Pointwise Reranker}
We now perform supervised fine-tuning on a \textbf{Qwen-4B} to obtain \textbf{zerank-1} as well as a smaller \textbf{Qwen-1.7B} model to obtain \textbf{zerank-1-small}, using the same relevance scores generated from preceding steps for each. We use a standard MSE loss function to fit $R_{\text{pred}}$ (our model) to $R_{\text{point}}$; $$\mathcal{L} = \frac{1}{|B|}\sum_{(q, d) \in B}{(R_{\text{pred}}(q, d)-R_{\text{point}}(q, d))^2} \; .$$ The result is our first-training-pass pointwise reranker $R'_{\text{point}}(q, d)$.

\subsection{Improving the Pairwise Reranker (RLHF)}

We improve on the pointwise model's performance through a variant of negative mining: we add additional data to the original (ensemble-inferenced) $N_Q$ pairwise scores used to train the pairwise reranker, based on failures of the pointwise reranker.

Specifically, for each query $q$ in our overall dataset, we let $d_{\text{human}}$ denote the document with the highest human-annotated score. Let $r_{\text{human}}$ be the rank of this document when we rank each document of $q$'s corpus $C$ using $R'_{\text{point}}$. For every sub-dataset constituent of our training dataset (id est, each constituent HuggingFace dataset), we manually set acceptable threshold rank values $t$. We determine this through factors such as how many documents are human-annotated for each query, how many human annotators were sampled for a given document-query pair, etc. If $r_{\text{human}} > t$, we consider this a failure. Consider now the document $d'$ that was ranked at position $r_{\text{human}}-1$ by the reranker. 

We will now inference the pairwise ensemble on $(d_{\text{human}}, d')$; we observe typically that the ensemble strongly prefers the human top-scored document, though evidently we ranked it worse. We add this $_{P}R_{\text{pair}}(q, d_{\text{human}}, d')$ to our pairwise reranker training dataset, while retaining the sole original example for that query, $_{P}R_{\text{pair}}(q, d_i, d_j)$, given by random sampling. 

We then retrain the pairwise reranker on the expanded dataset, and repeat all previous steps to obtain Elo scores with which to train a second iteration of our pointwise reranker on. \textbf{This gives us our final models, \textit{zerank-1} and \textit{zerank-small}.} 
\section{Results}
We conduct evaluations across a variety of specialized domains, and \textbf{zerank-1} consistently outperforms other state-of-the-art rerankers by significant margins. Our evaluation pipeline is open source and publicly available on \href{https://github.com/zeroentropy-ai/evals}{github}. Detailed results are on the subsequent page.

Specifically, we test zerank-1 and zerank-1-small against Cohere's \textit{rerank-v3.5}, Voyage AI's \textit{rerank-2}, and Salesforce's \textit{Llama-rank-v1} on a suite of public datasets with standard NDCG@10 metric (Table~\ref{tab:main-results}). We also evaluate said models across smaller private customer datasets to test for generalization and overfitting on eval datasets (Table~\ref{tab:results-2}). Notably, \textbf{neither zerank model was trained on any portion of the datasets evaluated}; in-fact, as recalled, the zELO training method does not use any human annotations.

\textbf{As can be seen, both zerank-1 and zerank-1-small offer large margins of improvement over existing SOTA methods, and these margins improve when tested on private datasets, indicating high generalization and wide applicability.}

Furthermore, zerank-1-small maintains much of the superior performance of zerank-1 over existing models despite being less than half the size. zerank-1-small is also available under a fully open Apache 2.0 license.

\begin{figure*}[!htb]
    \centering
    \includegraphics[width=1\linewidth]{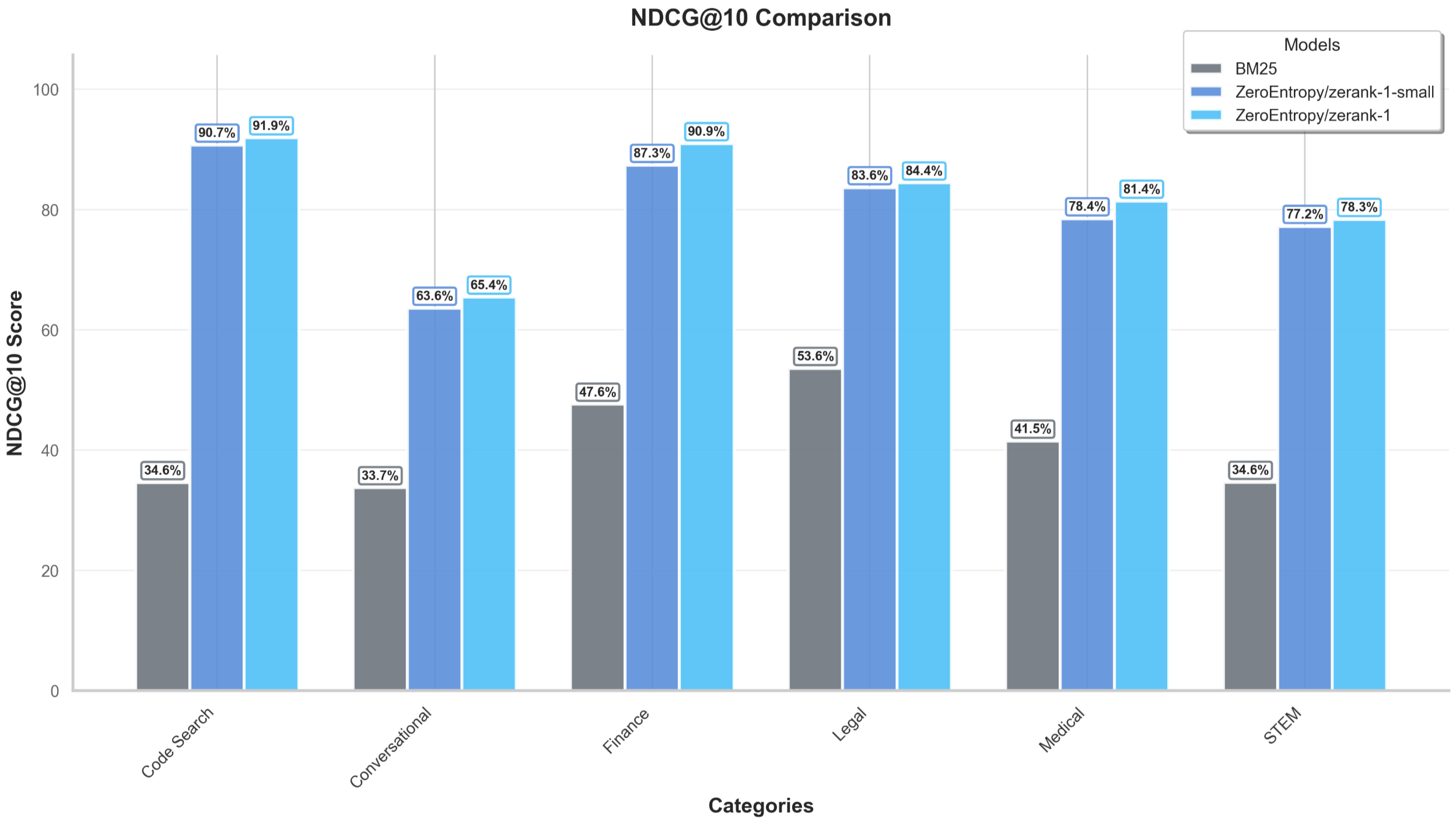}
    \includegraphics[width=1\linewidth]{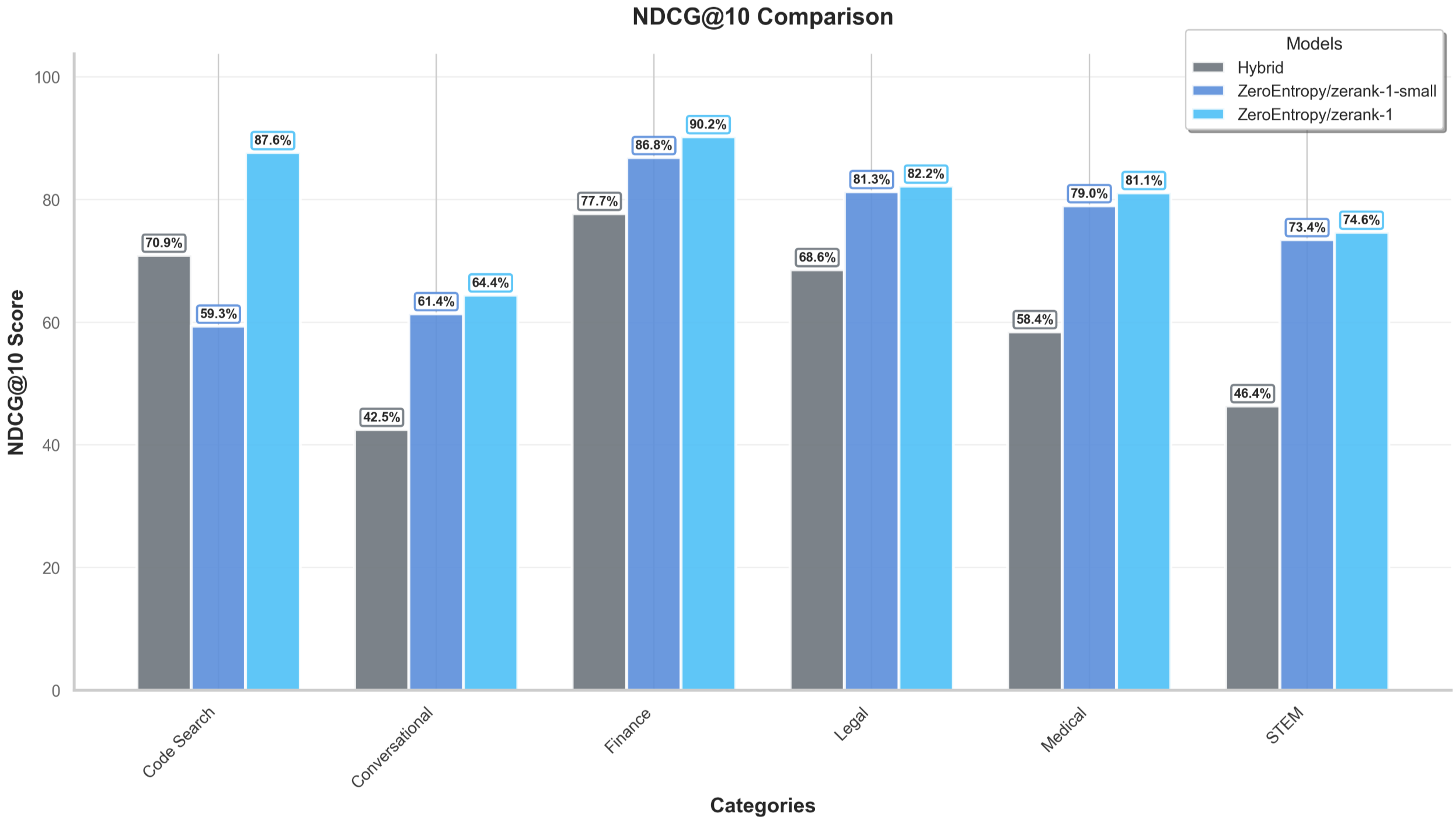}
    \caption{The upper graph displays NDCG@10 of BM25 alone, compared to using zerank-1 and zerank-1-small on the top 100 results of BM25. The lower graph is the same but for hybrid search instead of BM25.}
    \label{fig:hybridvsbm25}
\end{figure*}

Our evaluations use an initial retrieval with $k = 100$, and we test \textit{OpenAI-text-embedding-3-small}, BM25, and hybrid search. On all three initial retrieval methods, we find that our rerankers still significantly improve the NDCG@10 compared to initial retrieval (see Figure~\ref{fig:hybridvsbm25}).

Moreover, we achieve this performance while not compromising on speed:
\begin{table}[H]
\centering
\scalebox{0.8}{%
\begin{tabular}{|l|c|c|c|}
\hline
\textbf{Model} & \textbf{NDCG@10} & \textbf{Latency (12 KB)} & \textbf{Latency (150 KB)} \\
\hline
Jina  m0 & 0.7279 & 547.14 ± 66.84 ms & 2,543.8 ± 2,984.9 ms \\
\hline
Cohere 3.5 & 0.7091 & 171.5 ± 106.8 ms & 459.2 ± 87.9 ms \\
\hline
zerank-1 & \textbf{0.7683} & \textbf{149.7 ± 53.1 ms} & \textbf{314.4 ± 94.6 ms} \\
\hline
\end{tabular}
}
\caption{Zerank-1 offers improvement on SOTA NDCG@10 recall while being as fast or faster than competing models}
\label{tab:latency-comparison}
\end{table}

Lastly, we compare the use of zerank-1 against simply prompting a cheaper frontier LLM, \textbf{Gemini-v2.5-flash} (Gemini Team 2025), to perform pairwise comparisons using the same prompt as that used to generate pairwise reranker training data. We use the pairwise comparisons to generate Elo scores using Section 5.3's methods to obtain Gemini rankings (which we call gemini-reranker). We observe that our trained reranker still results in dramatically superior performance:
\begin{figure}[H]
    \centering
    \includegraphics[width=1\linewidth]{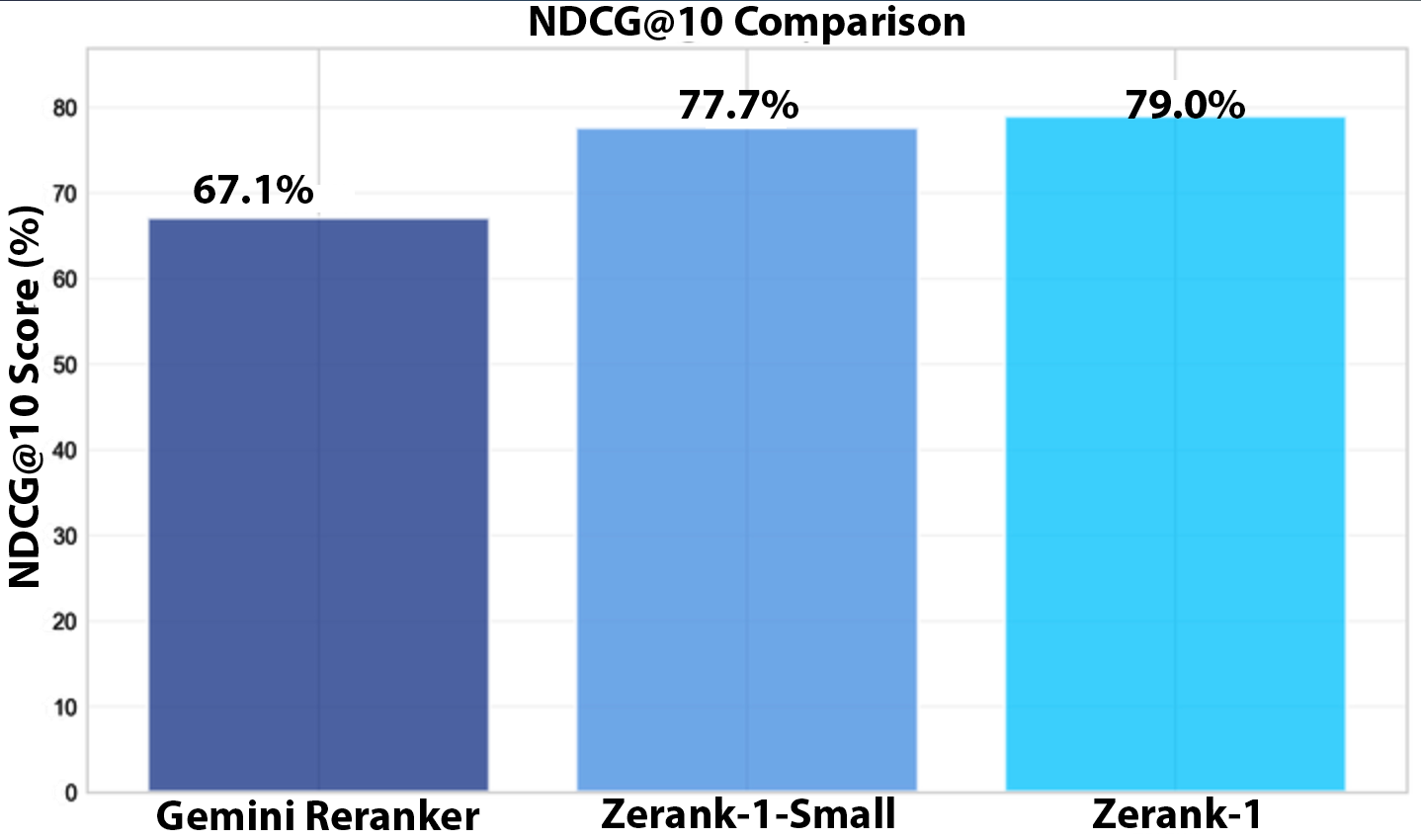}
    \label{fig:enter-label}
\end{figure}

\section{Conclusion}
This technical report introduces \textbf{zerank-1} and \textbf{zerank-1-small}, the new state-of-the-art models for reranking and information retrieval tasks. These rerankers were trained via a novel pipeline that largely rejects pointwise human annotations, and instead focuses on mathematically modeling query-document relevance scores using Elo-inspired scoring based on more meaningful pairwise comparisons. These models exhibit great cross-domain versatility, exhibiting strong performance on fields from code to finance to medicine, and we offer the weights of \href{https://huggingface.co/zeroentropy
}{both models} for non-commercial uses. 
\begin{figure*}[!htb]
    \centering
    \includegraphics[width=\linewidth]{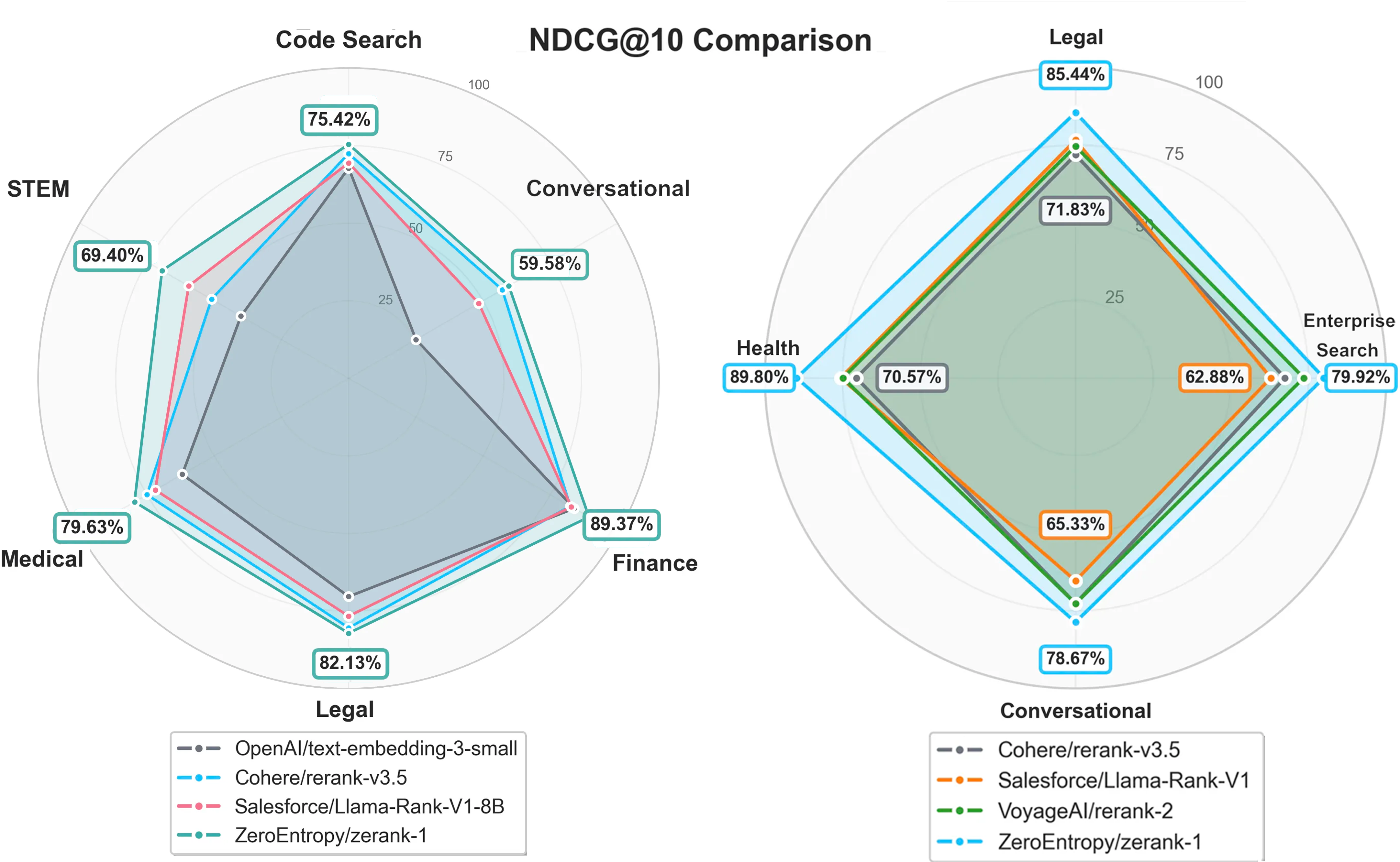}
    \caption{Caption}
    \label{fig:placeholder}
\end{figure*}
\begin{table*}[!htb]
\centering
\resizebox{\textwidth}{!}{%
\begin{tabular}{|| c ||c c c c c||} 
 \hline
 Task/Benchmark & Default(embedding) & cohere-rerank-v3.5 & Salesforce/Llama-rank-v1 & zerank-1-small & zerank-1 \\ [0.5ex] 
 \hline\hline
 Code & 0.678 &	0.724 &	0.694 & 0.730 &	\textbf{0.754} \\ 
 \hline
Conversational	&			
0.250 &	0.571 & 0.484 &	0.556 &	\textbf{0.596} \\
\hline				
Finance	&		
0.839 & 0.824	& 0.828 & 0.861	& \textbf{0.894} \\
\hline		
Legal &					
0.703 & 0.804	& 0.767 & 0.817 & \textbf{0.821} \\
\hline			
Medical	&		
0.619 &	0.750 &	0.719 & 0.773	& \textbf{0.796} \\
\hline
STEM &			
0.401 & 0.510	& 0.595 &	0.680 & \textbf{0.694} \\
\hline
\end{tabular}
}
\caption{Performance comparison across different tasks and benchmarks}
\label{tab:main-results}
\end{table*}
\begin{table*}[!htb]
\centering
\resizebox{\textwidth}{!}{%
\begin{tabular}{|| c ||c c c c c||} 
 \hline
 Task & Cohere/rerank-v3.5 & Salesforce/Llama-rank-v1 & VoyageAI/rerank-2 & zerank-1-small & zerank-1 \\ [0.5ex] 
 \hline\hline
 Legal & 0.718 &	0.766 &	0.746 & 0.799  &	\textbf{0.854} \\ 
 \hline
Enterprise Search	&			
0.674 &	0.629 & 0.735 &	 0.765 &	\textbf{0.799} \\
\hline				
Conversational	&		
0.727 & 0.653	& 0.727 & 	0.747 & \textbf{0.787} \\
\hline		
Healthcare &					
0.706 & 0.756	& 0.749 &  0.885 & \textbf{0.898} \\
\hline			
\end{tabular}
}
\caption{Performance comparison across different tasks and benchmarks}
\label{tab:results-2}
\end{table*}
\clearpage
\section{References}
\begin{enumerate}
    \item Yang, A., Li, A., Yang, B., Zhang, B., Hui, B., Zheng, B., Yu, B., Gao, C., Huang, C., Lv, C., Zheng, C., Liu, D., Zhou, F., Huang, F., Hu, F., Ge, H., Wei, H., Lin, H., Tang, J., … Zhang, Z. (2025, May 14). Qwen3 technical report (arXiv:2505.09388) [Preprint]. arXiv. \url{https://doi.org/10.48550/arXiv.2505.09388}
    \item Robertson, S. E., Walker, S., Jones, S., Hancock-Beaulieu, M. M., \& Gatford, M. (1995). Okapi at TREC-3. In D. K. Harman (Ed.), Proceedings of the Third Text REtrieval Conference (TREC-3) (pp. 109–126). National Institute of Standards and Technology (NIST). \url{https://trec.nist.gov/pubs/trec3/papers/city.ps.gz}    
    \item Cormack, G. V., Clarke, C. L. A., \& Buettcher, S. (2009). Reciprocal rank fusion outperforms Condorcet and individual rank learning methods. In Proceedings of the 32nd International ACM SIGIR Conference on Research and Development in Information Retrieval (pp. 758–759). ACM. 
    \url{https://plg.uwaterloo.ca/~gvcormac/cormacksigir09-rrf.pdf}
    \item Nogueira, R., \& Cho, K. (2020). Passage re‑ranking with BERT (arXiv:1901.04085v5) [Preprint]. arXiv. \url{https://doi.org/10.48550/arXiv.1901.04085}
    \item Together AI. (2024, June 6). Introducing Together Rerank API and Salesforce LlamaRank: Advancing reranking in retrieval. Together AI Blog. \url{https://www.together.ai/blog/together-rerank-api-and-salesforce-llamarank}
    \item Sean Lee, R., Huang, R., Shakir, A., \& Lipp, J. (2025, March 13). Baked‑in brilliance: Reranking meets RL with mxbai‑rerank‑v2. Mixedbread. \url{https://www.mixedbread.com/blog/mxbai-rerank-v2}
    \item Nogueira, R., Yang, W., Cho, K., \& Lin, J. (2019, October 31). Multi-stage document ranking with BERT (arXiv:1910.14424) [Preprint]. arXiv. \url{https://doi.org/10.48550/arXiv.1910.14424}
    \item OpenAI. (2024, January 25). New embedding models and API updates: text‑embedding‑3‑small \& text‑embedding‑3‑large. OpenAI. \url{https://openai.com/index/new-embedding-models-and-api-updates}
    \item Elo, A. E. (1978). The rating of chessplayers, past and present (Chapter 1). New York: Arco Publishing.
    \item Bradley, R. A., \& Terry, M. E. (1952). Rank analysis of incomplete block designs: I. The method of paired comparisons. Biometrika, 39(3–4), 324–345. \url{https://doi.org/10.2307/2334029}
    \item Zermelo, E. (1929). Die Berechnung der Turnier-Ergebnisse als ein Maximumproblem der Wahrscheinlichkeitsrechnung. Mathematische Zeitschrift, 29(1), 436–460. \url{https://doi.org/10.1007/BF01180541} 
    \item Bollobás, B. (2001). Random graphs (2nd ed.). Cambridge University Press. 
    \item Gemini Team. (2025, June 17). Gemini 2.5: Pushing the frontier with advanced reasoning, multimodality, long context, and next generation agentic capabilities. Google DeepMind. \url{https://storage.googleapis.com/deepmind-media/gemini/gemini_v2_5_report.pdf}

\end{enumerate}

\section{Appendix}

\subsection{Ensemble Inference Prompts}

\subsubsection{Single-query pairwise scores}
\label{sec:pairwiseprompt}
\textit{Task}

You are a relevance scoring system. Given a query and two documents (A and B), your job is to decide which document is more relevant to the given query. You should think carefully, considering the pros and cons between each document. For your first few sentences, consider the pros and cons of Document A. Then, spend some time thinking about Document B. Then, at the end, compare, and make a decision as to which one is more relevant. Do NOT make a decision in the beginning of your thoughts, stay open-minded until the last 1-2 sentences of your thoughts.

\textit{Scoring}

The score should range from -1.0 to 1.0, where negative means document A is more relevant, and positive means Document B is more relevant. You can pick any number from -1.0 to 1.0.

\bigskip

Note that here, whether $d_i$ or $d_j$ is document A is randomly chosen to mitigate any preference the model may have for either document position. We negate the score in that situation.

\end{document}